\documentclass{article}

\usepackage[utf8]{inputenc} 
\usepackage[T1]{fontenc}    
\usepackage{url}            
\usepackage{booktabs}       
\usepackage{amsfonts}       
\usepackage{nicefrac}       
\usepackage{microtype}      
\usepackage{graphicx}
\usepackage{subcaption}

\usepackage[preprint, nonatbib]{neurips_2020}

\usepackage{algorithm} 
\usepackage{algpseudocode} 
\usepackage{appendix}

\usepackage{amsmath}

\usepackage{cite}

\title{Policy Gradient from Demonstration and Curiosity}

\author{%
  Jie Chen\thanks{Jie Chen received his Ph.D degree from \emph{The University of Hong Kong} in 2017, and worked as a postdoctoral research fellow in \emph{Harvard University}. Currently, he is a senior researcher in Tencent.} \\
  Interactive Entertainment Group \\
  Tencent \\
  Shenzhen, China \\
  \texttt{jeremyjchen@tencent.com} \\
   \And
   Wenjun Xu \\
   Robotics Research Center \\
   Pengcheng Laboratory \\
   Shenzhen, China \\
   \texttt{xuwj@pcl.ac.cn} \\
}

\begin{document}

\maketitle

\begin{abstract}
  With reinforcement learning, an agent could learn complex behaviors from high-level abstractions of the task. However, exploration and reward shaping remained challenging for existing methods, especially in scenarios where the extrinsic feedback was sparse. Expert demonstrations have been investigated to solve these difficulties, but a tremendous number of high-quality demonstrations were usually required. In this work, an integrated policy gradient algorithm was proposed to boost exploration and facilitate intrinsic reward learning from only limited number of demonstrations. We achieved this by reformulating the original reward function with two additional terms, where the first term measured the Jensen-Shannon divergence between current policy and the expert's demonstrations, and the second term estimated the agent's uncertainty about the environment. The presented algorithm was evaluated by a range of simulated tasks with sparse extrinsic reward signals, where only one single demonstrated trajectory was provided to each task. Superior exploration efficiency and high average return were demonstrated in all tasks. Furthermore, it was found that the agent could imitate the expert's behavior and meanwhile sustain high return. 
\end{abstract}

\section{Introduction}

Over the last decade, reinforcement learning (RL) \cite{rl-intro} has achieved impressive success in various applications. Based on experiences collected through interaction with the environment, an agent learned a decision making strategy by means of trial and error. Mnih et al., 2015 \cite{dqn} trained an agent with Deep Q Networks (DQN) to play Atari games and achieved professional human level performance across a set of 49 games. In 2016, by incorporating human knowledge, Monte-Carlo Tree Search (MCTS), and self-play, Silver et al., \cite{alphago} built the very first agent, AlphaGo, to defeat a professional human Go player. Recently, with the help of large-scale distributed training infrastructure, reinforcement learning has been applied to real-time strategy multiplayer video games, which were thought to be very challenging due to issues like long time horizons, partially observable environment, and high-dimensional state and action spaces. The OpenAI Five (Berner et al., 2019) defeated the Dota 2 world champion in 2019 \cite{openai-five}. Vinyals et al., 2019 proposed the AlphaStar agent to master the game of StarCraft II and was rated at Grandmaster level \cite{alphastar}.

Alongside the tremendous success of reinforcement learning, exploration \cite{exploration} and reward shaping \cite{reward} remained challenging for existing algorithms. The agent struggled to learn especially when the extrinsic reward signals were sparse or the exploration spaces were huge. Recently, reinforcement learning from demonstration has attracted intensive research interest as a promising way to address these problems, however, existing algorithms usually required a tremendous number of high-quality demonstrations or included a human expert in the learning loop, which were often difficult or unavailable.

To this end, an integrated algorithm has been proposed in this work, named Policy Gradient from Demonstration and Curiosity (PGfDC), with the aim of facilitating exploration boosting and intrinsic reward learning from limited number of demonstrations in scenarios where the extrinsic reward signals were extremely sparse. The intuition behind PGfDC was: during interaction with the environment, when the extrinsic reward signals were sparse or even absent, an agent should imitate the demonstrated behaviors, when it got struggled in states where neither extrinsic reward nor demonstration data were available, an agent should attempt to explore novel states to minimize its uncertainty about the environment. After sufficient number of iterations, the agent could explore the environment on its own. 

To facilitate PGfDC, the original extrinsic reward function was reformulated by two additional terms which were derived from demonstration and curiosity, respectively. The demonstration term was established by computing the Jensen-Shannon divergence \cite{js} between the agent’s current policy and that of the expert. The concept of occupancy measure was introduced to approximate the policy divergence, by measuring the difference between self-generated data and the expert demonstration. To estimate the curiosity term, a neural network has been implemented to embed the agent’s observations and predict the consequences of its actions, uncertainties about the environment were measured to represent the curiosity reward. PGfDC was supposed to leverage expert demonstration and curiosity information to: (1) Reduce required number of demonstrations. (2) Improve exploration efficiency. (3) Imitate the expert and meanwhile achieve high return. These properties were desired by and could benefit real-world applications, for instance, human robot interaction, autonomous driving, and game AI. Furthermore, PGfDC was compatible with most policy gradient algorithms, e.g., Proximal Policy Optimization (PPO) \cite{ppo} and Trust Region Policy Optimization (TRPO) \cite{trpo}.In this work, PGfDC was evaluated on a range of grid world environments, where the original extrinsic reward signals were all extremely sparse.

\section{Related work}

\textbf{Curiosity driven exploration.} Various work focused on using curiosity to boost learning. Pathak et al., 2017 \cite{curiosity-pathak} designed an intrinsic curiosity module (ICM) by formulating curiosity reward as the uncertainty in an agent's ability to predict the consequence of its action. ICM improved exploration efficiency in scenarios where extrinsic reward signals were scarse or even absent. Burda et al., 2018 performed a large-scale study of purely curiosity-driven learning across 54 standard benchmark environments \cite{curiosity-burda}. However, purely curiosity-driven learning might sometimes become infeasible or dangerous in real-world settings. For example, in autonomous driving and human robot interaction, unexpected movements might occur and lead to catastrophe.

\textbf{Reinforcement learning from demonstration.} Expert demonstrations have been introduced to guide the learning process. Hester et al., 2017 proposed the Deep Q-learning from Demonstrations (DQfD) and stored the demonstrations in experience replay buffer \cite{dqfd}. In Silver et al., 2016 \cite{alphago}, demonstration data was used to pre-train the policy network. Although reinforcement learning from demonstration had the potential to relieve exploration dilemma in sparse reward scenarios, existing algorithms tended to require a tremendous number of high-quality data but often failed to fully leverage the value of the demonstrations.

\textbf{Inverse reinforcement learning.} Inverse reinforcement learning (IRL) and inverse optimal control (IOC) have provided a set of algorithms to directly learn the reward functions from demonstrations, as in Ng et al., 2000 \cite{irl-ng}, Abbeel et al., 2004 \cite{irl-abbeel}, Ziebart et al., 2008 \cite{ziebart}, and Finn et al., 2016 \cite{gcl}. However, it was difficult to make an IRL algorithm effective since: (1) IRL asked for a number of high-quality expert demonstrations. (2) IRL was inherently underdefined as different reward functions might result in similar behaviors.

\textbf{Reward learning from preference.} A large amount of work have been conducted on reinforcement learning from human preferences or ratings. Christiano et al., 2017 explored learning objectives defined in terms of human preferences between pairs of trajectory segments, and demonstrated the effectiveness of the method on Atari games and simulated robot locomotion without access to the extrinsic reward signals \cite{christiano}. In Ibarz et al., 2018 \cite{ibarz}, expert demonstrations and trajectory preferences were combined, where a reward function was learned from the preferences and the demonstrations were used by a DQfD algorithm. However, preference learning might get struggled when encountered with tasks where qualified experts were not available. Moreover, the number and the quality of preferences required by an agent grew with the complexity of environments, making the learning process inefficient and sometimes even intractable.

\section{Preliminaries}

\subsection{Markov decision process}

In this work, the problems considered were under the standard Markov Decision Process (MDP) setting. An MDP was formalized by the tuple: $\langle \mathcal{S}, \mathcal{A}, r, p_0, T, \gamma \rangle$, where $\mathcal{S}$ and $\mathcal{A}$ represented for the state space and action space, $r = r(s, a, s')$ was the reward function, $p_0$ was the probability distribution of the initial state, $T = T(s'|s, a)$ denoted the transition function of the environment, and $\gamma\in(0, 1)$ was the discount factor. An agent interacted with the environment over time based on policy $\pi(a|s)$, mapping state to action probability. At time step $t$, the agent received $s_t$ from the state space $\mathcal{S}$, selected $a_t$ from the action space $\mathcal{A}$ according to $\pi(a_t|s_t)$, transitioned to the next state $s_{t+1}$ based on $T = T(s_{t+1}|s_t, a_t)$, and received a scalar reward signal $r_t = r(s_t, a_t, s_{t+1})$. The discounted return was $\mathcal{R}_t = \sum_{k=0}^{\infty} \gamma^k r_{t+k}$, and expectation of $\mathcal{R}_t$ was usually evaluated to reflect performance of the policy $\pi$:
\begin{equation}
    J(\pi) = \mathbb{E}_{\pi}[ r(s, a, s') ] = \mathbb{E}_{(s_0, a_0, s_1, a_1, s_2, ...)}[\mathcal{R}_t]
\end{equation}
where $(s_0, a_0, s_1, a_1, s_2, ...)$ was a trajectory generated from interaction with the environment. Correspondingly, the value function could be defined as $V_{\pi}(s) = \mathbb{E}_{\pi}[\mathcal{R}_t | s_t = s]$, the action value function was $Q_{\pi}(s, a) = \mathbb{E}_{\pi}[\mathcal{R}_t | s_t = s, a_t = a]$, and the advantage function was $A_{\pi}(s, a) = Q_{\pi}(s, a) - V_{\pi}(s)$. The objective of RL algorithms was to discover the optimal policy that can maximize the expectation of discounted return $\mathbb{E}_{\pi}[ \mathcal{R}_t ]$.

\subsection{Policy gradient}

Unlike value-based reinforcement learning, the policy gradient methods directly modelled and optimized the policy $\pi_{\theta} (a|s)$ parameterized by $\theta$. And the learning objective was defined as:
\begin{equation*}
    J(\pi_\theta) = \mathbb{E}_{\pi_\theta}[ r(s, a, s') ] = \sum_{s \in \mathcal{S}} d^{\pi_\theta}(s) V_{\pi_\theta}(s) = \sum_{s \in \mathcal{S}} d^{\pi_\theta}(s) \sum_{a \in \mathcal{A}} \pi_{\theta} (a|s) Q_{\pi_\theta}(s, a)
\end{equation*}
 where $J(\pi_\theta)$ could be used to measure the performance of policy $\pi_{\theta} (a|s)$, where $d^{\pi_\theta}(s)$ represented the stationary distribution of Markov chain for $\pi_{\theta} (a|s)$. According to the policy gradient theorem:
 \begin{equation*}
     \nabla_{\theta} J(\pi_\theta) = \sum_{s \in \mathcal{S}} d^{\pi_\theta}(s) \sum_{a \in \mathcal{A}} \nabla_{\theta} \pi_{\theta} (a|s) Q_{\pi_\theta}(s, a) = \mathbb{E}_{\pi_\theta}[\nabla_{\theta} log \pi_{\theta} (a|s) Q_{\pi_\theta}(s, a)]
 \end{equation*}
 where $\theta$ could be optimized via gradient ascent. To solve $\nabla_{\theta} J(\pi_\theta)$, $Q_{\pi_\theta}(s, a)$ should be computed. Normally, $Q_{\pi_\theta}(s, a)$ could be approximated with methods like Monte-Carlo estimation (REINFORCE), Temporal-Difference learning, or with an auxiliary critic model (actor-critic policy gradient). Furthermore, to reduce variance, the advantage function $A_{\pi_\theta}(s, a)$ was introduced to substitute $Q_{\pi_\theta}(s, a)$, and hence, $\nabla_{\theta} J(\pi_\theta) = \mathbb{E}_{\pi_\theta}[\nabla_{\theta} log \pi_{\theta} (a|s) A_{\pi_\theta}(s, a)]$.

\section{Methodology}
\label{headings}

With the widespread use and advances of RL, the significance and difficulty of exploration and reward design have been highlighted. In real-world scenarios, the extrinsic reward signal was usually extremely sparse and hard to be reshaped, which affected the exploration efficiency. Introducing demonstrations or curiosity has proven to be effective in sparse reward settings. The demonstrations were often exploited in the following ways: (1) Store in the experience replay buffer. (2) Pre-train the policy network. (3) Infer an intrinsic reward function. Curiosity was deployed to encourage the agent to explore novel states or perform actions to reduce its uncertainty about the environment dynamics. In this work, to fully leverage demonstration data and curiosity, the above two ideas were combined to formulate a new policy gradient method which was boosted from both demonstration and curiosity (PGfDC). PGfDC was supposed to outperform existing methods since: (1) It required limited number of demonstrations. (2) It could guarantee superior exploration efficiency. (3) It could imitate the expert and meanwhile achieve high return, which was desired in areas like human robot interaction, autonomous driving, and game AI.

\begin{figure}[t!]
  \centering
  \includegraphics[width=\linewidth]{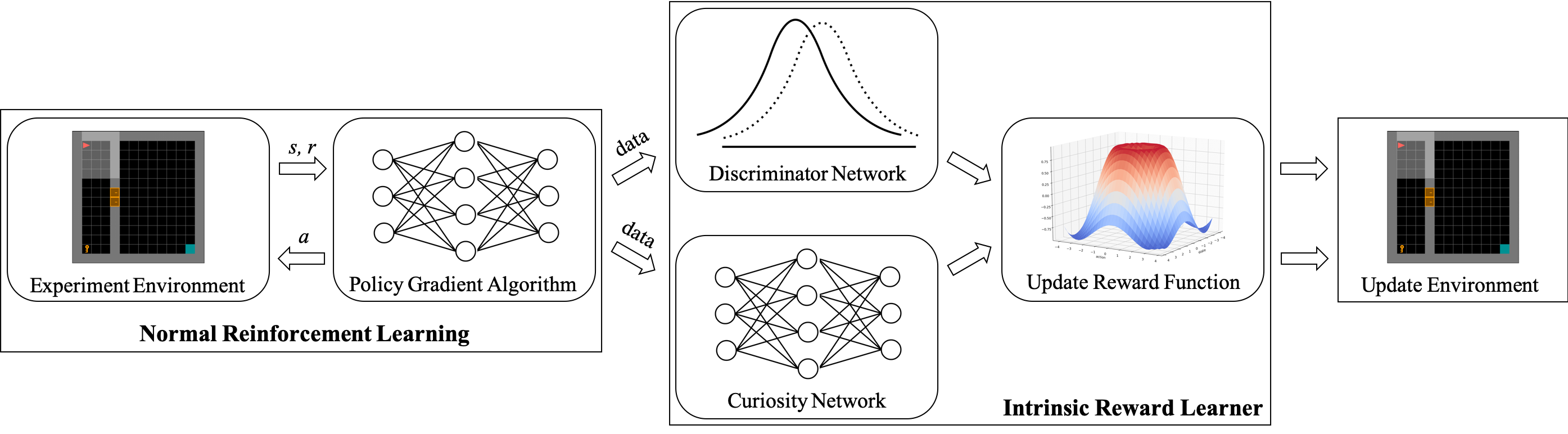}
  \caption{Workflow of the PGfDC algorithm.}
  \label{fig:workflow}
\end{figure}

The overall workflow of the proposed PGfDC algorithm was shown in Figure. \ref{fig:workflow}. There were two sub-modules, namely the normal RL module and the intrinsic reward learner. For the reinforcement learning module, the agent interacted with the environment and received reward signals estimated by the current reward function $\widetilde{r}_k$:
\begin{equation}
    \widetilde{r}^k = r_{e} + \lambda_d r_d^k + \lambda_c r_c^k
\end{equation}
where $r^{e}$ was the original extrinsic reward function of the environment, $r_d^k$ represented the intrinsic reward function learned from demonstrations at the $k^{th}$ iteration, $r_c^k$ was the intrinsic reward function learned from curiosity at the $k^{th}$ iteration, and $\lambda_d$ and $\lambda_c$ were the corresponding weighting coefficients. The collected interaction data was stored as $\langle s, a, r, s' \rangle$, and was sent to the intrinsic reward learner. Within the intrinsic reward learner, the discriminator network was updated with pre-stored expert demonstrations and the interaction data, and the curiosity network was simultaneously optimized with gradients computed from the collected interaction data. Then the reward function was updated to $\widetilde{r}^{k+1}$. The intrinsic reward learner could work synchronously or asynchronously \cite{async} with the standard reinforcement learning module. Details of PGfDC were summarized in the appendix.

\subsection{Policy gradient from demonstration and curiosity}

\subsubsection{Reward learning from demonstration}

Reinforcement learning from demonstration has proved to be an efficient and intuitive way of transferring expert's knowledge and preference to the agent. The agent could either infer a reward function from the demonstrations as in inverse reinforcement learning, or boost its exploration through a pre-trained policy. However, existing methods usually asked for a tremendous number of high-quality demonstration data while failed to fully leverage the data. To address these issues, the demonstrations were used to formulate an additional penalty term to the original learning objective in this work, measuring the Jensen-Shannon divergence between the current policy $\pi_{\theta} (a|s)$ and the demonstrations. Specifically, suppose the provided demonstrations were expressed as $\mathcal{D}^E = \langle \tau_0, \tau_1, ..., \tau_N \rangle$, where $\tau_i = \langle (s_0^i, a_0^i), (s_1^i, a_1^i), ..., (s_T^i, a_T^i) \rangle$, and $\mathcal{D}^E$ was generated from an implicit expert policy $\pi_E$. Then the reformulated learning objective was obtained:
\begin{equation}
    \min_{\theta} \mathcal{L}(\pi_\theta) = -J(\pi_\theta) + \lambda_d \mathcal{D}_{JS} (\pi_\theta, \pi_E)
\end{equation}
where $\lambda_d \in (0, 1)$ was the weighting coefficient. It was impossible to directly estimate $\mathcal{D}_{JS} (\pi_\theta, \pi_E)$ as $\pi_E$ was unknown, thus the concept of occupancy measure was introduced to approximate $\mathcal{D}_{JS} (\pi_\theta, \pi_E)$. 

\textbf{Definition 1.} (\emph{Occupancy measure}) Let $\rho_{\pi_\theta}(s): \mathcal{S} \rightarrow \mathbb{R}$ denote the unnormalized distribution of state visitation by following policy $\pi_\theta$ in the environment, $\rho_{\pi_\theta}(s) = \sum_{t=0}^{\infty}\gamma^t P(s_t=s|\pi_\theta)$, then the unnormalized distribution of state-action pairs $\rho_{\pi_\theta}(s, a) = \rho_{\pi_\theta}(s) \pi_\theta(a|s)$ was termed occupancy measure of policy $\pi_\theta$. 

According to Theorem 2 of (Syed et al., 2008 \cite{syed}), $\pi_\theta$ was the only policy whose occupancy measure was $\rho_{\pi_\theta}$, given that $\rho_{\pi_\theta}$ was the occupancy measure for $\pi_\theta(a|s) = \frac{\rho_{\pi_\theta}(s, a)} {\sum_{a'} \rho_{\pi_\theta}(s, a')}$. Therefore, the Jensen-Shannon divergence between $\pi_\theta$ and $\pi_E$ could be substituted by:
\begin{equation}
    \mathcal{D}_{JS} (\pi_\theta, \pi_E) = \mathcal{D}_{JS}  (\rho_{\pi_\theta}, \rho_{\pi_E})
\end{equation}
Kang et al., 2018 \cite{pofd} derived a lower bound for $\mathcal{D}_{JS}  (\rho_{\pi_\theta}, \rho_{\pi_E})$, which could be reformulated as:
\begin{equation}
    \mathcal{D}_{JS}  (\rho_{\pi_\theta}, \rho_{\pi_E}) \geq \max_w \mathbb{E}_{(s, a) \sim \rho_{\pi_E}}[log(D_w(s, a))] + \mathbb{E}_{(s, a) \sim \rho_{\pi_\theta}}[1 - log D_w(s, a)]
\end{equation}
where $D_w(s, a) : \mathcal{S} \times \mathcal{A} \rightarrow (0, 1)$, and $w$ was the parameters. Actually, the right side of Equation (5) could be viewed as the learning objective of discriminator in Generative Adversarial Network (GAN), with $\pi_\theta$ working as the generator. In order to train $D_w(s, a)$, state-action pairs from $\rho_{\pi_E}$ were labeled as true, while the state-action pairs generated by $\rho_{\pi_\theta}$ were labeled as false. Substitute Equation (5) into Equation (3), the following learning objective was obtained:
\begin{equation}
    \min_\theta \max_w -J(\pi_\theta) + \lambda_d (\mathbb{E}_{(s, a) \sim \rho_{\pi_E}}[log D_w(s, a)] + \mathbb{E}_{(s, a) \sim \rho_{\pi_\theta}}[1 - log D_w(s, a)])
\end{equation}
which was equivalent to:
\begin{equation}
    \min_\theta \max_w -\mathbb{E}_{\pi_\theta}[ r(s, a, s') ] + \lambda_d \mathbb{E}_{\pi_E}[log D_w(s, a)] + \lambda_d \mathbb{E}_{\pi_\theta}[1 - log D_w(s, a)]
\end{equation}
Furthermore, Equation (7) could be re-organized as:
\begin{equation}
    \min_\theta \max_w -\mathbb{E}_{\pi_\theta}[ r(s, a, s') + \lambda_d (log D_w(s, a) - 1)] + \lambda_d \mathbb{E}_{\pi_E}[log D_w(s, a)]
\end{equation}
In Equation (8), the original reward function was reshaped by $\lambda_d (log D_w(s, a) - 1)$, as the constant $-1$ could be removed. Thus, based on Equation (1), provided with input tuple $\langle s, a, s' \rangle$, the demonstration reward was:
\begin{equation}
    r_d(s, a, s') = log D_w(s, a)
\end{equation}

\begin{figure}[t!]
  \centering
  \includegraphics[width=\linewidth]{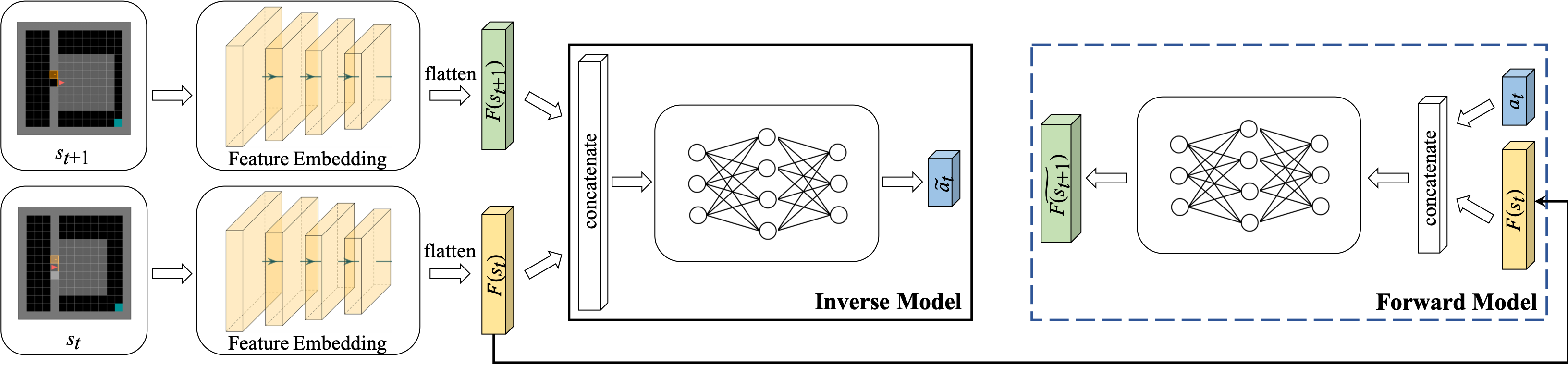}
  \caption{Reward learning from curiosity.}
  \label{fig:curiosity}
\end{figure}

\subsubsection{Reward learning from curiosity}

Following \cite{curiosity-pathak}, curiosity in PGfDC consisted of three sub-modules: feature embedding $G_e$, inverse model $G_i$, and the forward model $G_f$. In $G_e$, the input state $s \in \mathcal{S}$ was encoded as a feature vector $F(s)$. Then the feature vectors of two consequent states, $F(s_t)$ and $F(s_{t+1})$ were concatenated and fed into $G_i$ to generate prediction for the action $\widetilde{a}_t$ taken by the agent to move from $s_t$ to $s_{t+1}$. $G_e$ and $G_i$ could be combined to formulate a joint model:
\begin{equation}
    G_{ei} = G_{ei} (\widetilde{a}_t | s_t, s_{t+1}, \theta_{ei})
\end{equation}
where $\theta_{ei}$ was the network parameters and was optimized through minimizing $\mathcal{L}_{ei} (\widetilde{a}_t, a_t)$. As discrete actions were used in this work, $\mathcal{L}_{ei}$ could be cross-entropy. For the forward model $G_f$, feature vector $F(s_t)$ and the corresponding action $a_t$ were taken as the input to predict feature vector $\widetilde{F}(s_{t+1})$ of the state at next time step:
\begin{equation}
    G_f = G_f (\widetilde{F}(s_{t+1}) | F(s_t), a_t, \theta_f)
\end{equation}
where the network parameters $\theta_f$ were optimized by minimizing the mean squared loss function $\mathcal{L}_f (\widetilde{F}(s_{t+1}), F(s_{t+1})) = \frac{1}{2} \| \widetilde{F}(s_{t+1}) - F(s_{t+1}) \|_2^2$. In this work, $\theta_{ei}$ and $\theta_f$ were jointly updated and the loss functions $\mathcal{L}_{ei}$ and $\mathcal{L}_f$ were combined and formulated as:
\begin{equation}
    \min_{\theta_{ei}, \theta_f} \mathcal{L}_{curiosity} = (1 - \beta) \mathcal{L}_{ei} + \beta \mathcal{L}_f
\end{equation}
where $\beta$ was the controlling weighting factor and $\beta \in (0, 1)$. The training data was collected while the agent was interacting with the environment and was stored in the tuple $\langle s_t, a_t, s_{t+1} \rangle$. $\mathcal{L}_f$ was used to calculate the curiosity reward, and a transformation function $\frac{e^z - 1}{e^z + 1}$ was applied to $\mathcal{L}_f$ to scale it to the range of $[0, 1]$. Therefore, given the input tuple $\langle s, a, s' \rangle$, the curiosity reward was:
\begin{equation}
    r_c(s, a, s') = \frac{e^{\mathcal{L}_f(\widetilde{F}(s'), F(s'))} - 1}{e^{\mathcal{L}_f(\widetilde{F}(s'), F(s'))} + 1}
\end{equation}
Figure. \ref{fig:curiosity} illustrated the workflow of the curiosity reward module.
 
\section{Experimental Evaluation}
\label{others}

In this section, performance of the proposed PGfDC algorithm was experimentally evaluated on the following aspects: (1) Given limited number of demonstrations, could PGfDC guarantee superior exploration efficiency? (2) Considering the extremely sparse extrinsic reward of environments, could PGfDC guarantee high return at convergence? (3) Given demonstrations from the expert, could PGfDC imitate the expert’s behavioral preference and meanwhile achieve high empirical return?

\begin{figure}
  \centering
  \begin{subfigure}[b]{0.3\linewidth}
    \includegraphics[width=\linewidth]{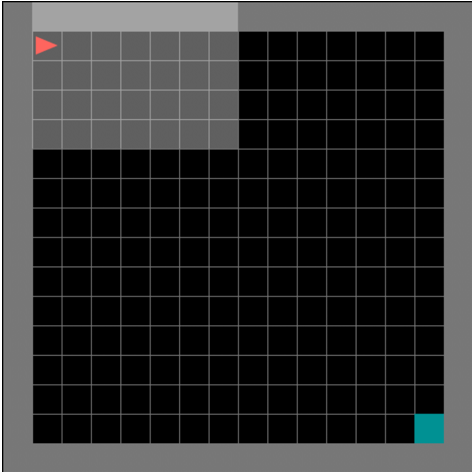}
    \caption{}
  \end{subfigure}
  \begin{subfigure}[b]{0.3\linewidth}
    \includegraphics[width=\linewidth]{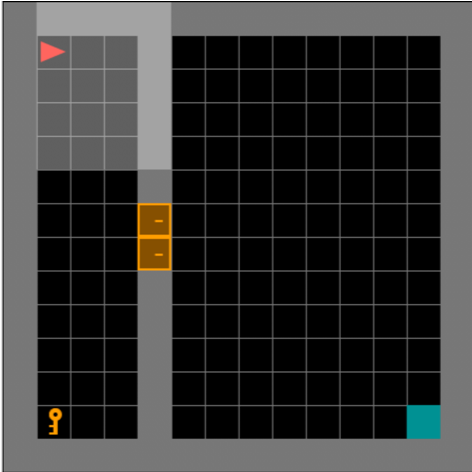}
    \caption{}
  \end{subfigure}
  \begin{subfigure}[b]{0.3\linewidth}
    \includegraphics[width=\linewidth]{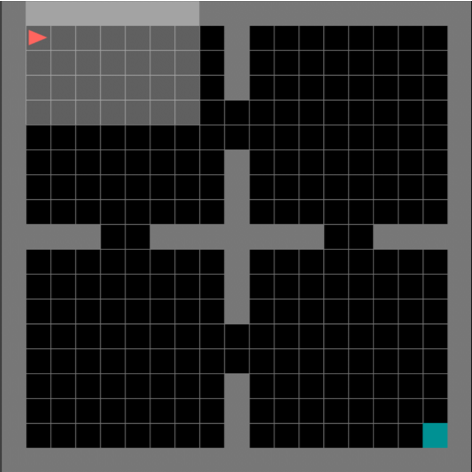}
    \caption{}
  \end{subfigure}
  \caption{Experiment environments. (a) $14\times14-Grid World$. (b) $14\times14-Key World$. (c) $4-Room Maze$.}
  \label{fig:environment}
\end{figure}

\subsection{Environment settings}

In order to comprehensively investigate performance of the proposed algorithm, three variants of the grid world environment have been designed and implemented \cite{minigrid}: (1) An empty grid world with the size of $14\times14$, where the agent was required to reach for the terminal state ($14\times14-Grid World$, Figure. \ref{fig:environment}(a)). (2) A grid world environment of the size $14\times14$, where the agent had to pick up a key first, and then open a door with the key to reach for the terminal state ($14\times14-Key World$, Figure. \ref{fig:environment}(b)). (3) A grid world maze composed of four connected rooms, where each of the room has the size of $8\times8$, the agent was required to navigate through the rooms and reach for the terminal state ($4-Room Maze$, Figure. \ref{fig:environment}(c)). For all of the three environments, a sparse extrinsic reward was given when the agent reached the terminal state:
\begin{equation}
    r_e = 1 - 0.9\frac{n_t}{N_{max}}
\end{equation}
where $n_t$ was the number of time steps taken by the agent, and $N_{max}$ denoted the maximum number of time steps. For each environment, only one single demonstrated trajectory was provided to the agent: $\tau = \{s_t, a_t\}_{t=0}^T$, and the PGfDC algorithm was compared with four baselines: (1) A human expert. (2) Policy trained with Advantage Actor Critic (A2C) \cite{a3c}. (3) Policy learned with Proximal Policy Optimization (PPO). (4) A random policy. In the following subsections, implementation details of PGfDC were briefly overviewed, including the policy network, the discriminator, and the curiosity.

\subsection{Network architectures}

\subsubsection{Policy network}

A2C, PPO, and PGfDC shared the same policy network architecture, where the input state $s_t$ was passed through three consecutive convolution layers, with filter numbers of 16, 32, and 64, respectively, and the kernel size was $2\times2$. A rectified linear unit (RELU) was used after each convolution layer, and a max pooling operation with the size of $2\times2$ was deployed after the first convolution layer. The output of the last convolution layer was flattened to be fed into two separate fully connected layers to predict the action probability distribution and the value function, where each fully connected layer had the size of 64 and a Tanh activation function after it. To get the action probability distribution, another fully connected layer with size equivalent as that of the action space was implemented, followed by a softmax operation. On the other hand, an output layer with 1 hidden unit was used to predict the value function.

\subsubsection{Discriminator network}

The input action $a_t$ was passed through a 2-layer MLP with RELU activations and 16 and 8 hidden units correspondingly to obtain the action feature vector. The input state $s_t$ was passed through a sequence of two convolution layers with filter numbers of 16 and 32, a RELU and a max pooling operation with size of $2\times2$ were deployed after each convolution layer. The output of the last convolution layer was flattened and fed into a 2-layer MLP with RELU activations and 16 and 4 hidden units correspondingly to obtain the state feature vector. The action feature vector and the state feature vector were concatenated and passed through a fully connected layer with size of 4, followed by a RELU. To predict the discriminator reward $r_d$, an output layer with 1 hidden unit and a sigmoid activation function was used. Learning rate of the discriminator was set to be $10^{-3}$ for all of the three environments.

\begin{figure}
  \centering
  \begin{subfigure}[b]{0.328\linewidth}
    \includegraphics[width=\linewidth]{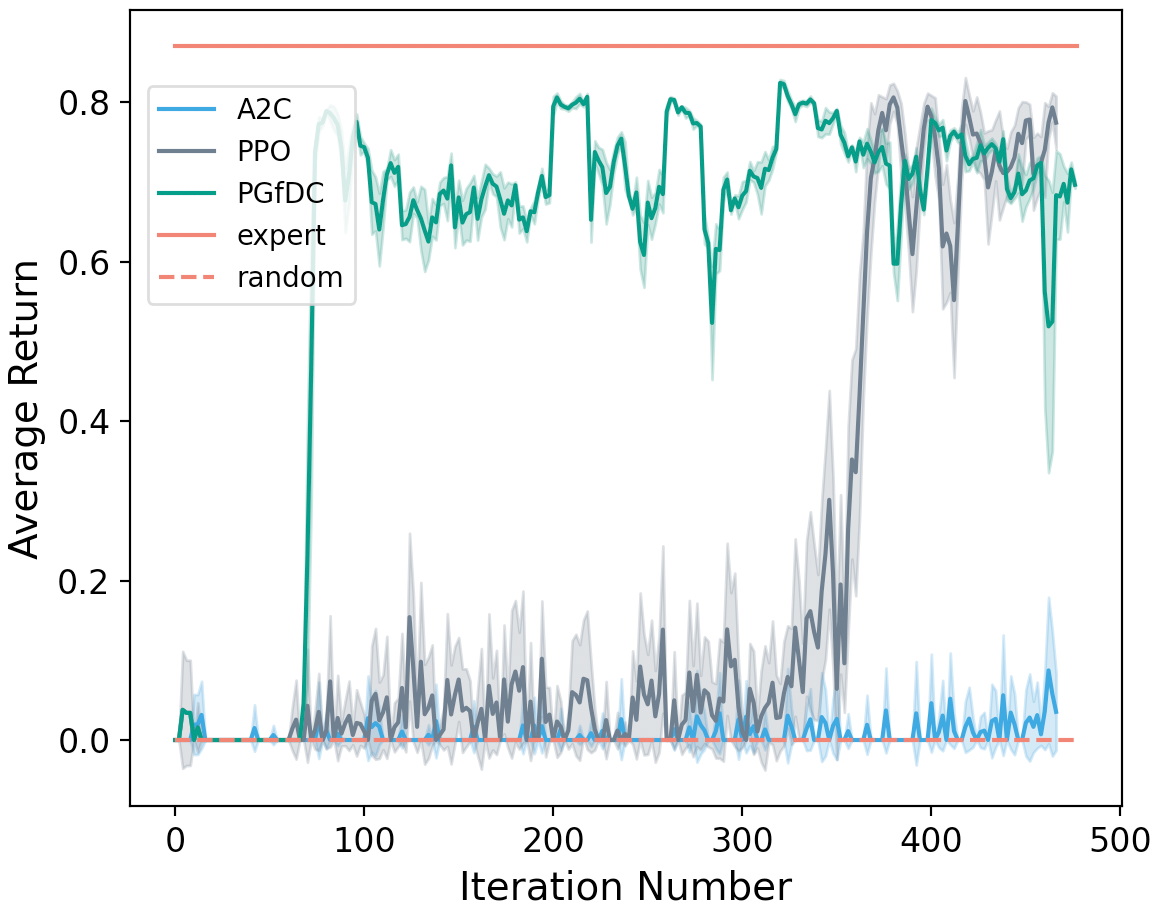}
    \caption{}
  \end{subfigure}
  \begin{subfigure}[b]{0.328\linewidth}
    \includegraphics[width=\linewidth]{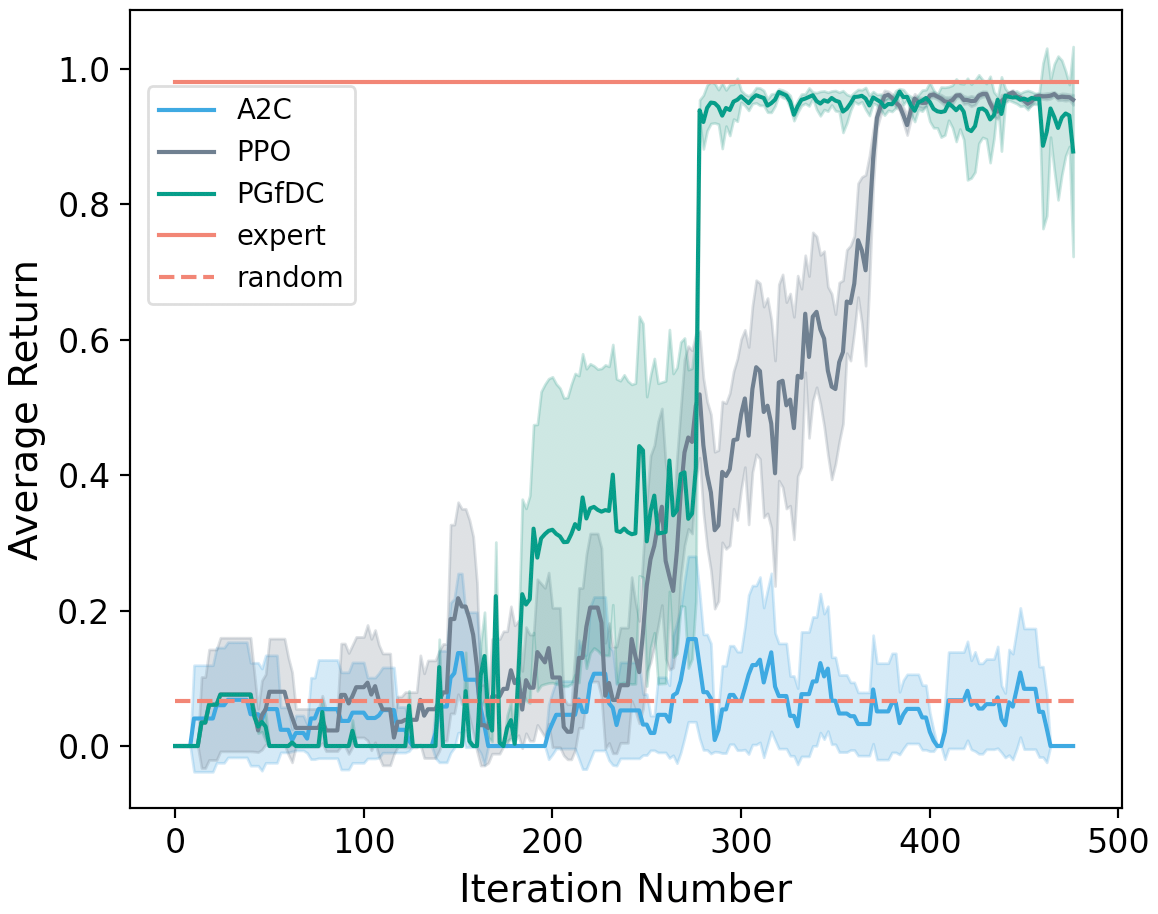}
    \caption{}
  \end{subfigure}
  \begin{subfigure}[b]{0.328\linewidth}
    \includegraphics[width=\linewidth]{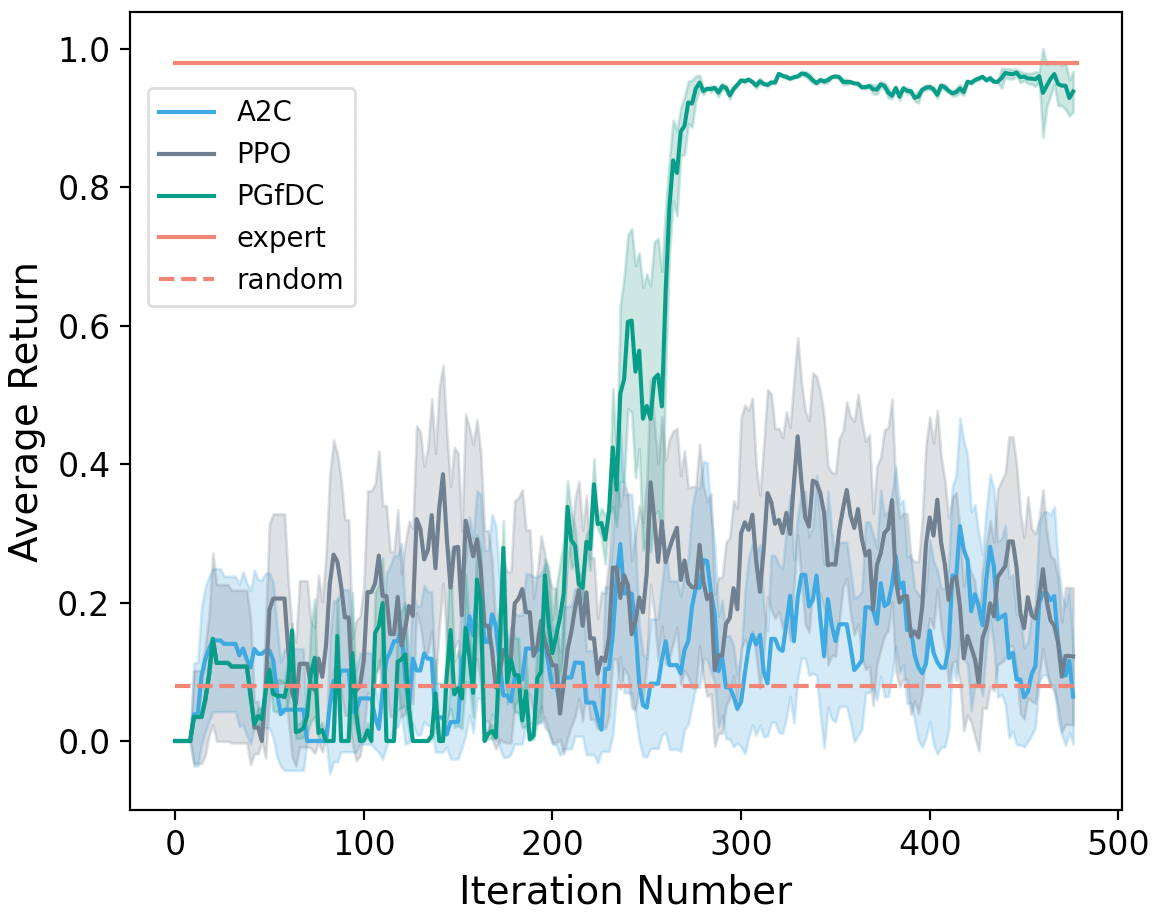}
    \caption{}
  \end{subfigure}
  \caption{Learning curves of PGfDC against baselines. (a) $14\times14-Grid World$. (b) $14\times14-Key World$. (c) $4-Room Maze$.}
  \label{fig:learning curve}
\end{figure}

\begin{figure}
  \centering
  \begin{subfigure}[b]{0.328\linewidth}
    \includegraphics[width=\linewidth]{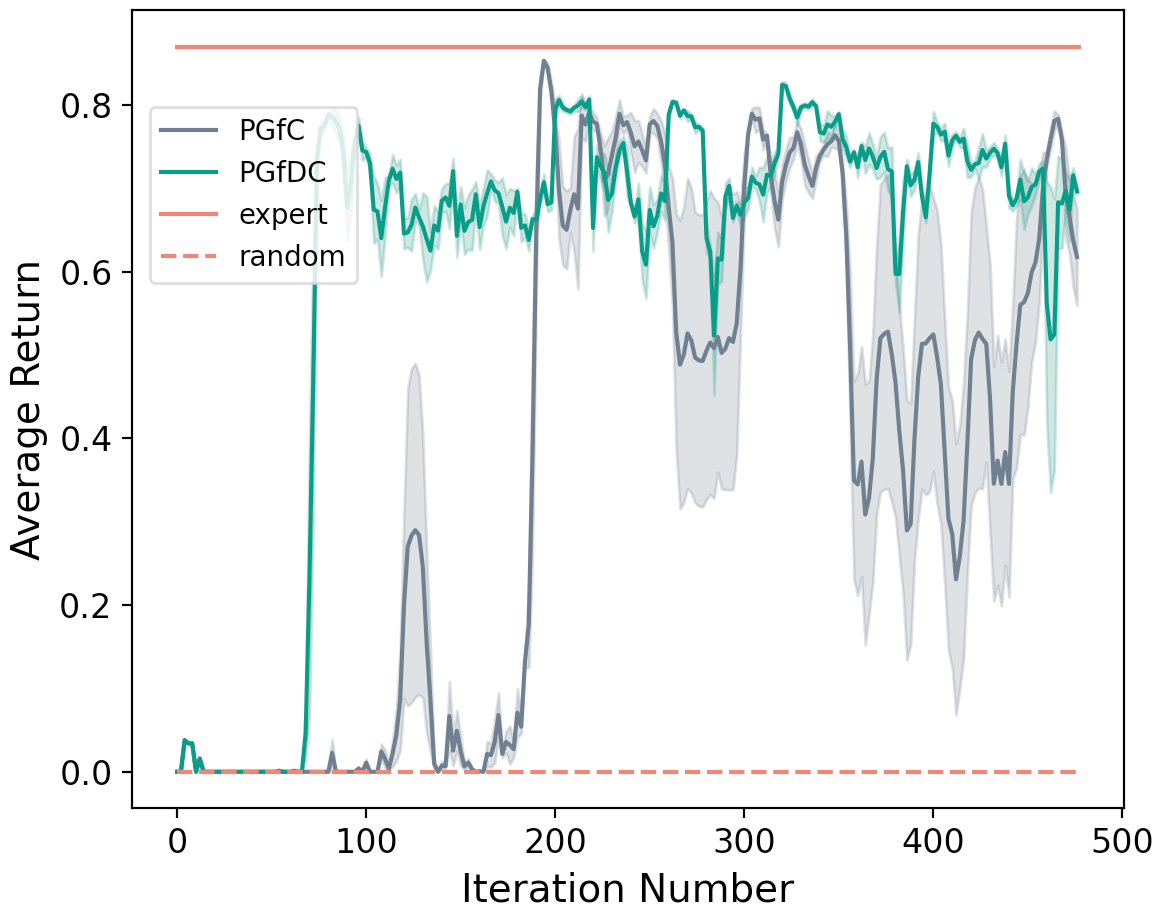}
    \caption{}
  \end{subfigure}
  \begin{subfigure}[b]{0.328\linewidth}
    \includegraphics[width=\linewidth]{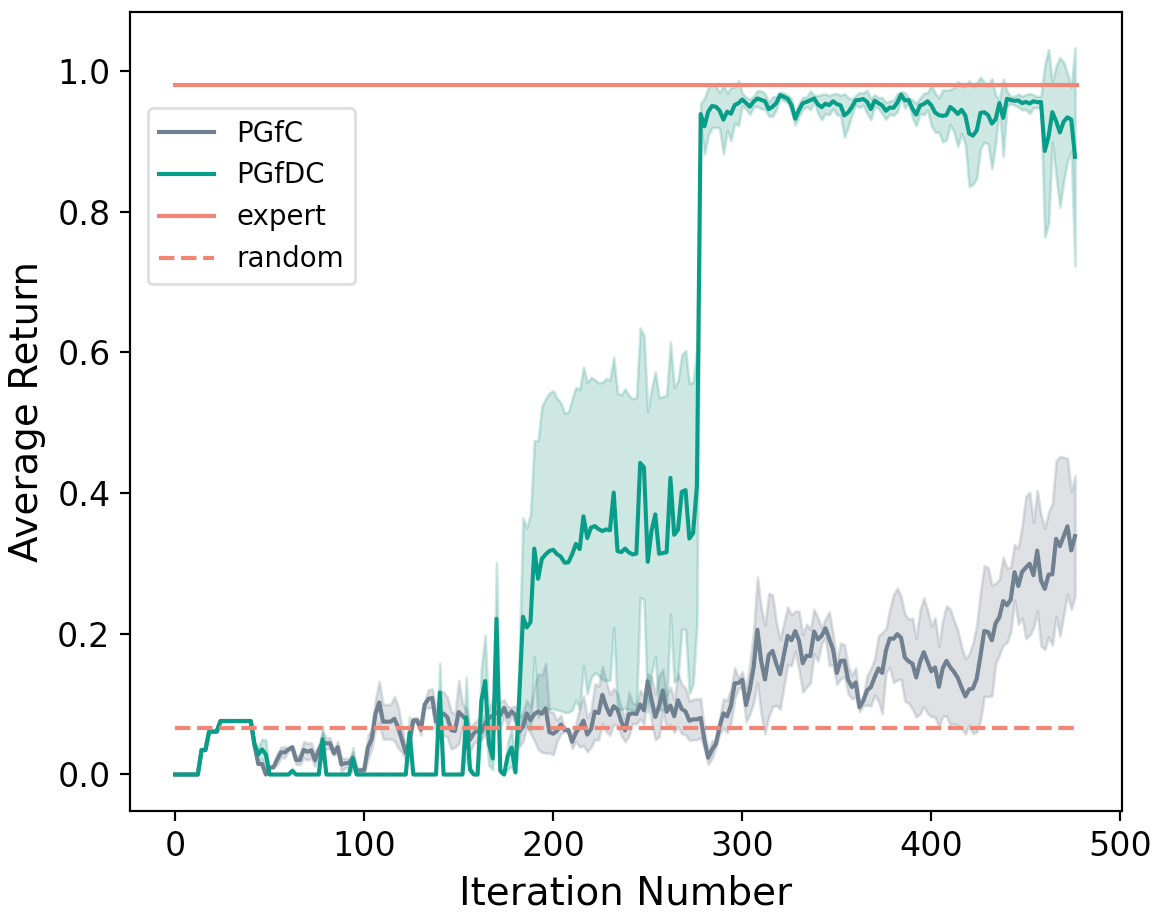}
    \caption{}
  \end{subfigure}
  \begin{subfigure}[b]{0.328\linewidth}
    \includegraphics[width=\linewidth]{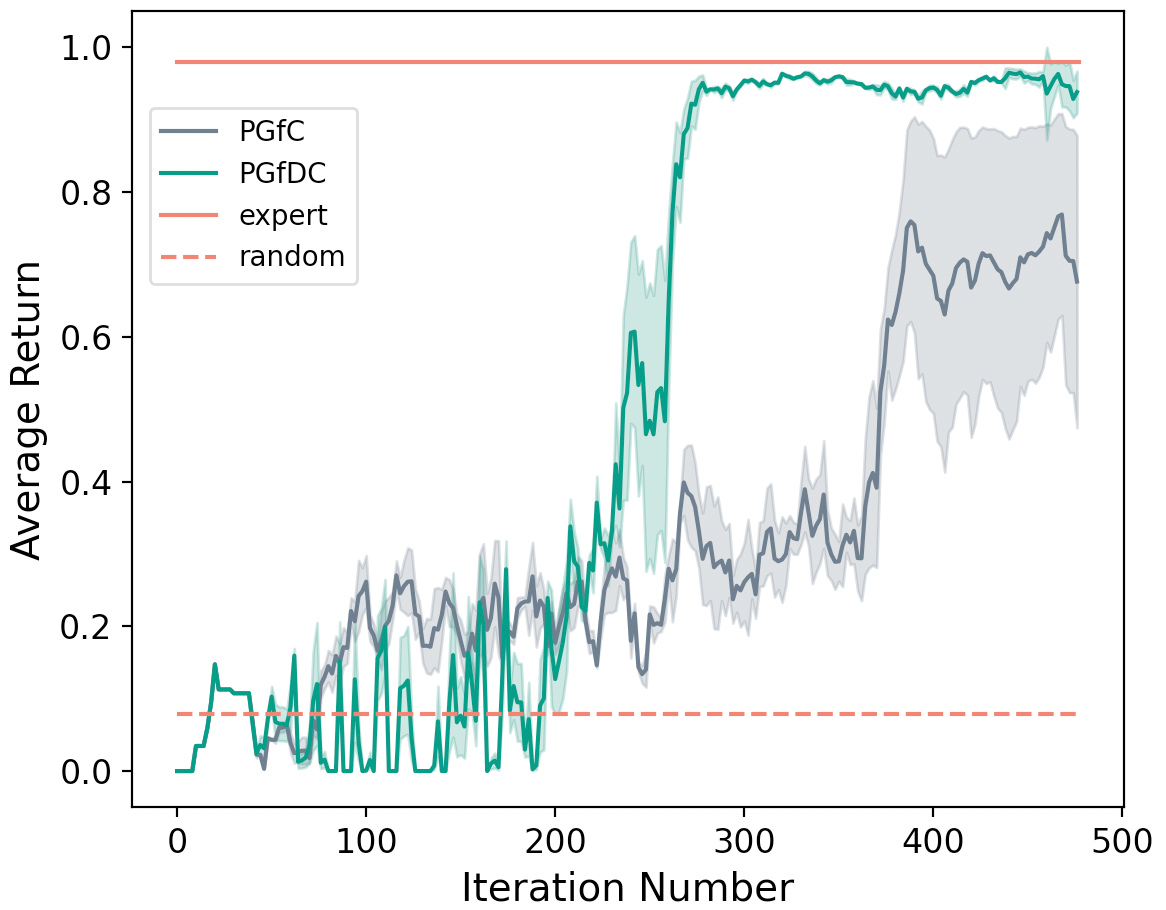}
    \caption{}
  \end{subfigure}
  \caption{Learning curves of PGfDC with and without demonstration information \cite{curiosity-pathak}. (a) $14\times14-Grid World$. (b) $14\times14-Key World$. (c) $4-Room Maze$.}
  \label{fig:learning curve2}
\end{figure}

\subsubsection{Curiosity network}

Adapted from \cite{curiosity-pathak}, architecture of the curiosity network was illustrated in Figure. \ref{fig:curiosity}. The curiosity module was composed of three components: the feature embedding $G_e$, the inverse model $G_i$, and the forward model $G_f$. The feature embedding mapped the input states $s_t$ and $s_{t+1}$ into feature vectors $F(s_t)$ and $F(s_{t+1})$ with a sequence of four convolution layers with the same filter number of 16 and kernel size of $3\times3$, an ELU activation function was used after each convolution layer. The output of the last convolution layer was flattened to generate a 32-dimensional feature vector. For the inverse model, $F(s_t)$ and $F(s_{t+1})$ were concatenated and passed through a fully connected layer with RELU activation and 64 hidden units, followed by an output layer activated by the sigmoid function to predict the action. In the forward model, the embedded feature vector $F(s_t)$ and action $a_t$ were concatenated and fed into a fully connected layer with 128 hidden units and the RELU, followed by an output layer with 32 hidden units to predict the feature vector of $s_{t+1}$, $\widetilde{F}(s_{t+1})$. For all the environments, learning rate of the curiosity was set to be $10^{-3}$, and $\beta$ was $10^{-2}$.

\begin{figure}
  \centering
  \begin{subfigure}[b]{0.45\linewidth}
    \includegraphics[width=\linewidth]{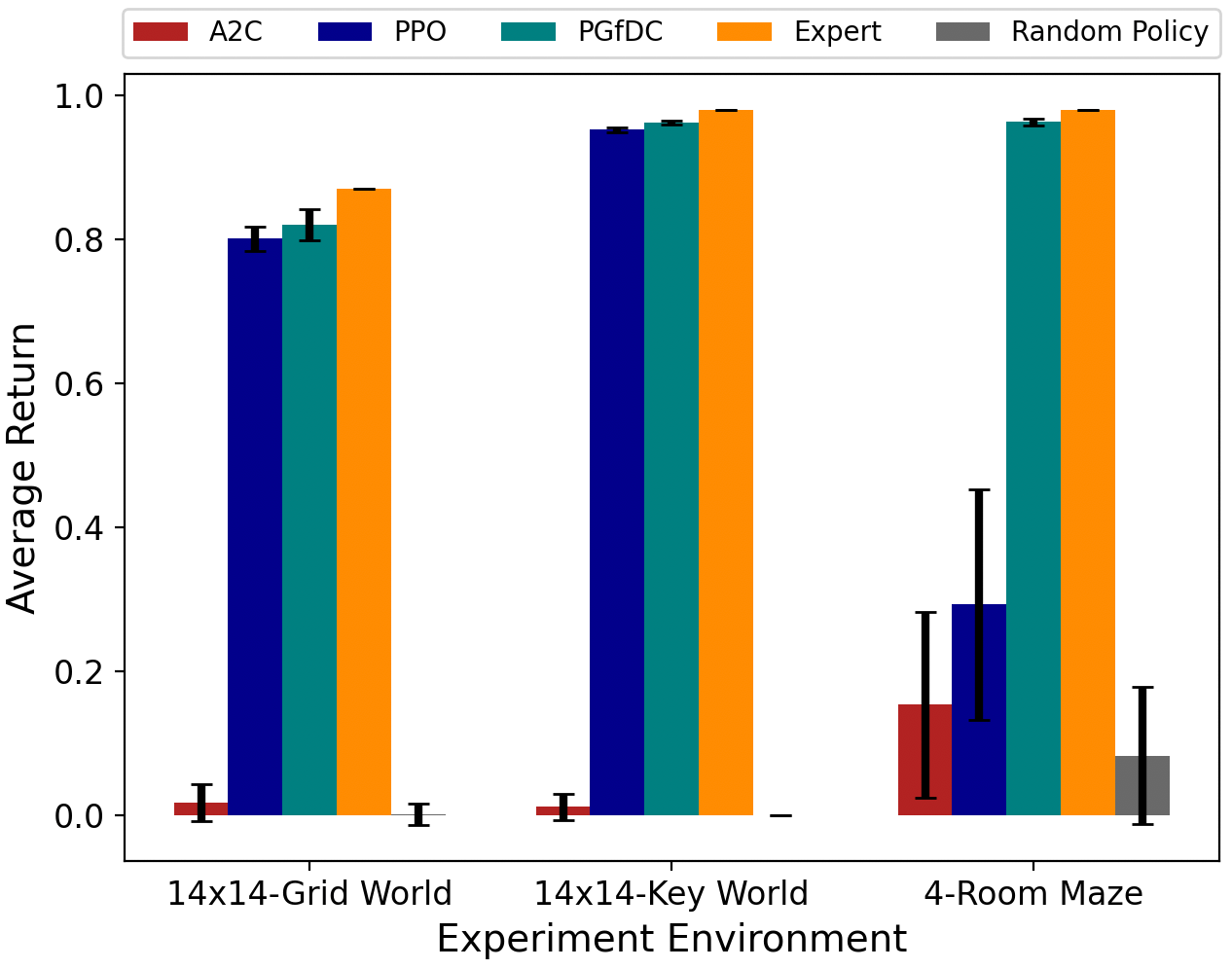}
    \caption{Average return}
  \end{subfigure}
  \begin{subfigure}[b]{0.45\linewidth}
    \includegraphics[width=\linewidth]{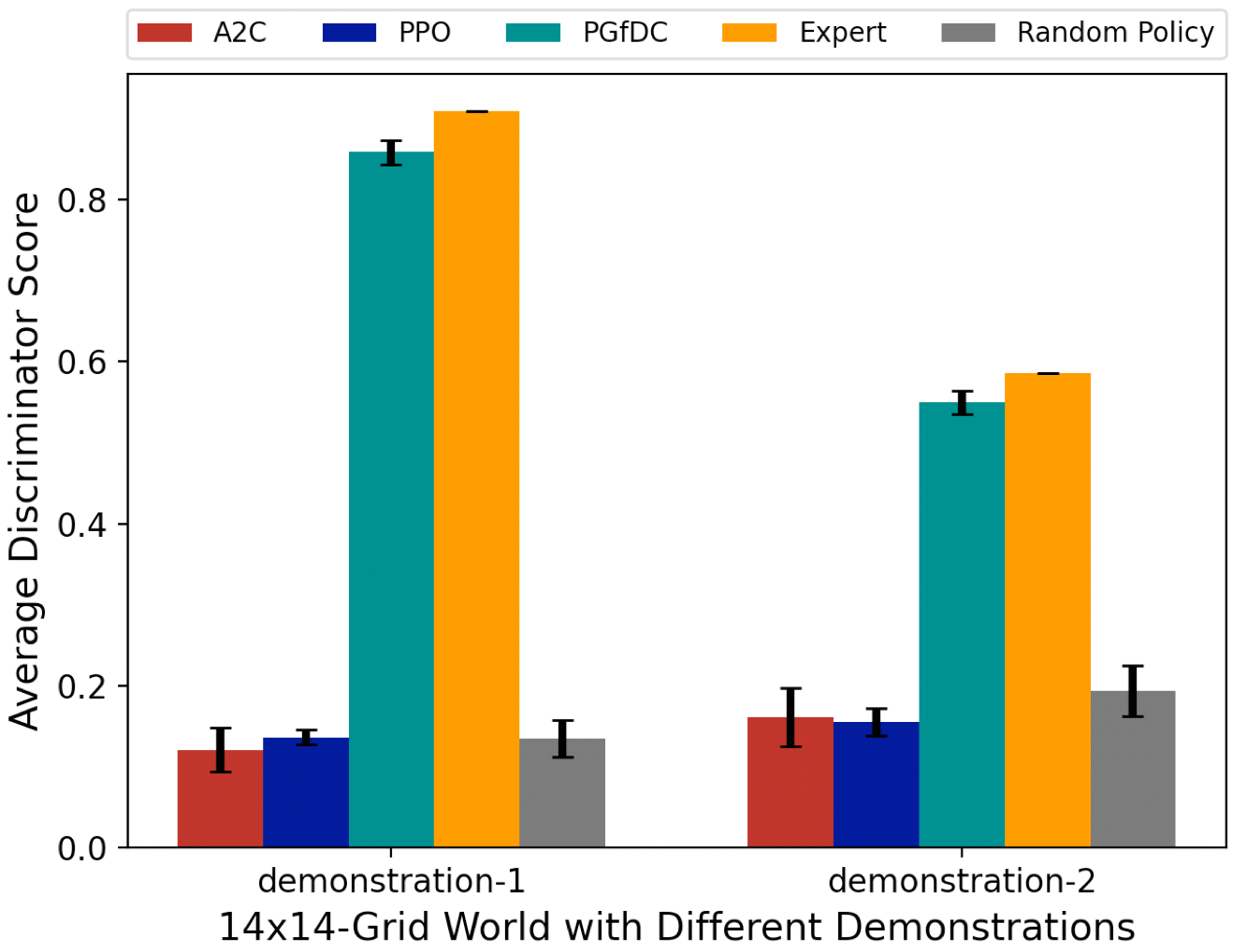}
    \caption{Average discriminator score}
  \end{subfigure}
  \caption{Evaluation results of the trained policies.}
  \label{fig:histogram}
\end{figure}

\subsection{Experimental results}

Figure. \ref{fig:environment} showed the three experimental environments, and the corresponding learning curves were depicted in Figure. \ref{fig:learning curve}. Only one single demonstrated trajectory was provided to the agent in each environment. The hyperparameters of PGfDC as well as other experimental details were summarized in the Appendix. Figure. \ref{fig:learning curve}(a) illustrated the learning curves of the environment $14\times14-Grid World$. The proposed PGfDC algorithm started to converge after about 100 iterations, the PPO converged after approximately 360 iterations, while the A2C algorithm failed to converge within 460 iterations. Figure. \ref{fig:learning curve}(b) showed the learning curves in the environment $14\times14-Key World$, where PGfDC succeeded to converge at around $280^{th}$ iteration, the PPO algorithm succeeded to converge after about 380 iterations, while A2C still failed within 460 iterations. The learning curves for the environment $4-Room Maze$ were provided in Figure. \ref{fig:learning curve}(c). In this task, the proposed PGfDC started to converge at $300^{th}$ iteration approximately, however, neither PPO nor A2C converged within 460 iterations in this task. In addition, an ablation study has been performed by removing the demonstration reward term $r_d$. Figure. \ref{fig:learning curve2} illustrated the comparison of PGfDC with and without demonstration information. Without demonstration, PGfDC degenerated to the method presented in \cite{curiosity-pathak}. In all the tasks, PGfDC with demonstration outperformed the one without demonstration.

Evaluation of the learned policies was performed using 10 different random seeds, and for each seed, the average return was computed with 10 independent rollout episodes. The evaluation results were shown in Figure. \ref{fig:histogram}(a). In all of the three experiments, the proposed PGfDC algorithm achieved higher returns compared with A2C and PPO. Specifically, the average returns of PGfDC were 0.820, 0.962, and 0.963 for the environments $14\times14-Grid World$, $14\times14-Key World$, and $4-Room Maze$, respectively. In the $4-Room Maze$ environment, neither A2C nor PPO succeeded. A2C achieved 0.154 on average, while the average return of PPO was 0.293.

The abovementioned experimental evaluations have validated that: (1) Given one single demonstrated trajectory, the proposed PGfDC algorithm succeeded to converge at much higher exploration efficiency compared with A2C and PPO in all three environments. (2) Considering the extremely sparse extrinsic reward of the tasks, PGfDC successfully achieved higher average return against A2C and PPO in all of the tasks. In order to investigate the third aspect of the proposed algorithm, which was: provided with demonstrations from the expert, could PGfDC imitate the expert’s behavior while achieve high return at the same time, the $14\times14-Grid World$ environment was used as the testbed. 

Two independent and distinct demonstrated trajectories were provided to the agent separately to facilitate two independent experimental runs, and thus two discriminators have been learned with the demonstrations to determine whether the input sample came from the expert or not (based on Equation (5)). The experimental results were shown in Figure. \ref{fig:histogram}(b), and the average discriminator score, $\mathbb{E}_{\pi_\theta}[D_w(s, a)]$, was calculated to reflect the similarity between expert and the policy. Again, the evaluations were performed using 10 different random seeds, and for each seed, the average discriminator score was computed with 10 independent rollout episodes. Provided with demonstration-1, PGfDC achieved 0.859 on average. As comparison, the discriminator scored the expert for 0.910, while A2C and PPO only received 0.121 and 0.137, respectively, given that the random policy achieved an average discriminator score of 0.135. With demonstration-2, the proposed PGfDC algorithm achieved an average score of 0.550, the expert’s average score was 0.586, while A2C, PPO and the random policy received 0.162, 0.156, and 0.194, respectively. Therefore, the experimental results have validated that the proposed algorithm had the potential of imitating expert and meanwhile achieving considerably high return.

\section{Conclusion}

Alongside the development of reinforcement learning algorithms, reward shaping and exploration remained challenging for existing methods. An agent might struggle to discover useful information, especially when interacting with environment where the extrinsic feedback was sparse. An integrated algorithm has been developed in this work, the PGfDC, with the purpose to boost exploration and facilitate intrinsic reward learning in sparse reward scenarios from only limited number of demonstrations. In PGfDC, the original reward function was reformulated by two additional terms, $r_d$ and $r_c$, where $r_d$ was the intrinsic reward learned from demonstrations with a discriminator network, and $r_c$ represented the intrinsic reward signal derived from curiosity. To comprehensively evaluate the performance of PGfDC, three grid world-like environments have been designed, where the original extrinsic reward was extremely sparse. For each environment, only one single demonstrated trajectory was provided. The experimental results validated that: (1) Provided with limited number of demonstrations, PGfDC could guarantee superior exploration efficiency. (2) PGfDC could achieve high return in sparse reward scenarios. (3) PGfDC could imitate the expert’s behavioral preference and meanwhile achieve high empirical return. Details of the implementation were given. And in theory, PGfDC was compatible with most policy gradient algorithms. Future work would be focused on extending PGfDC to real-world applications, e.g., computer games and human robot interaction.

\medskip

\small

\bibliography{ref}{}
\bibliographystyle{plain}

\appendix

\newpage
\section{Broader impact}

For real world sequential decision making problems, the proposed PGfDC algorithm has the potential to reduce the difficulty of exploration. For instance, the magnitude of states and actions involved in Go can reach to $10^{170}$ and $10^{360}$ \cite{silver2016mastering}, while in the Multi-player Online Battle Arena (MOBA) game, Honor of Kings, the magnitude of states and actions involved in the 1v1 mode can reach to $10^{600}$ and $10^{18000}$. To train one single hero in the Honor of Kings, Ye et al., \cite{ye2019mastering} used $48$ NVIDIA P40 GPU cards and $18,000$ CPU cores, not to mention that the game has dozens of heroes in total. The presented PGfDC algorithm demonstrates the potential to reduce the overall training cost of applying reinforcement learning (RL) to such large scale games. Moreover, for physical systems like robots and autonomous vehicles, it is impossible to build thousands of duplicates to exhaustively explore the state and action spaces as in computer games, which prevents the widespread application of RL. The PGfDC algorithm may benefit the deployment of RL to such areas. We have also experimentally validated that the proposed PGfDC algorithm can work properly with only limited number of demonstrations, which further paved its way to physical systems, where it is usually difficult to collect a large number of high-quality demonstrations.

On the other hand, the presented PGfDC algorithm shows the potential to reduce the burden of reward shaping. In many real world settings, the external reward signals are extremely sparse, which increases the difficulty of training a RL agent. Reward shaping becomes essential in such scenarios, since it can make the original reward signals dense and guide the agent to collect more effective training samples. For instance, the original feedback in Dota $2$ is provided only at the end of the game as win, lose, or draw. In order to train a RL agent to play Dota 2, Berner et al., \cite{berner2019dota} shaped the original reward function by introducing terms as hero death, health changed, lane assign, etc. However, it is non-trivial to do reward shaping, especially when the number of reward terms increases. The PGfDC algorithm serves as an alternative to automatically shape the original reward function with only limited number of demonstrations, which may save tremendous human efforts.

\newpage
\section{Hyperparameters}

\begin{table}[h]
  \caption{Hyperparameters used by PGfDC}
  \label{hyperparameters}
  \centering
  \begin{tabular}{llllllllllll}
    \toprule
    \multicolumn{6}{c}{PPO hyperparameters} \\
    \cmidrule(r){1-6}
    $\gamma$ & $\alpha$ & $\lambda_{GAE}$ & $\alpha_{entropy}$ & $\alpha_{value}$ & clip & $\lambda_c$ & $\lambda_d$ & $\mathcal{S}$ & $\mathcal{A}$ & $N_{max}$ \\
    \midrule
    0.99 & $10^{-3}$ & 0.95 & $10^{-2}$ & 0.5 & 0.2 & $10^{-3}$ & $10^{-2}$
    & $\mathbb{R}^{3\times7\times7}$ & $\mathbb{R}^3$ & 192 \\
    0.99 & $10^{-3}$ & 0.95 & $10^{-2}$ & 0.5 & 0.2 & $10^{-4}$ & $10^{-3}$
    & $\mathbb{R}^{3\times7\times7}$ & $\mathbb{R}^5$ & 1960 \\
    0.99 & $10^{-3}$ & 0.95 & $10^{-2}$ & 0.5 & 0.2 & $10^{-4}$ & $10^{-3}$
    & $\mathbb{R}^{3\times7\times7}$ & $\mathbb{R}^3$ & 1536 \\
    \bottomrule
  \end{tabular}
\end{table}

\newpage
\section{Synchronous PGfDC}

\begin{algorithm}
	\caption{Synchronous PGfDC with PPO} 
	\begin{algorithmic}[1]
	\State Initialize policy parameters $\theta_p^0$, value function parameters $\phi_p^0$, and clip $\epsilon$
	\State Initialize discriminator parameters $w^0$, curiosity parameters $\langle \theta_{ei}^0, \theta_{f}^0 \rangle$
	\State Initialize experience replay buffer $\mathcal{D}^G$ to store generated trajectories
	\State Load expert demonstrations $\mathcal{D}^E$, state-action pairs $\langle s, a \rangle$ in $\mathcal{D}^E$ were labeled as 1
		\For {$k=1,2,\ldots$}
		    \For {$l=1,2,\ldots$}
		        \State Collect set of trajectories $\mathcal{D}^l$ by running policy $\pi (\theta_p^l)$ in the environment
		        \State Compute state value estimates, $\widetilde{V}_{\pi (\theta_p^l)} (s_t)$
		        \State Compute advantage estimates, $\widetilde{A}_{\pi (\theta_p^l)} (s_t, a_t)$
		        \State Update $\theta_p^l$ by minimizing the PPO-Clip objective:
		        \begin{equation}
		            \theta_p^{l+1} = argmax_{\theta} \frac{1}{|\mathcal{D}^l|T} \sum_{\tau \in {\mathcal{D}^l}} \sum_{t=0}^T \min [\frac{\pi_{\theta} (s_t, a_t)}{\pi_{\theta_p^l} (s_t, a_t)} A^{\pi_{\theta_p^l}} (s_t, a_t), g \langle \epsilon, A^{\pi_{\theta_p^l}} (s_t, a_t) \rangle] \tag{A.1}
		        \end{equation}
		        \State Update $\phi_p^l$ by maximizing the mean-squared error:
		        \begin{equation}
		            \phi_p^{l+1} = argmin_{\phi} \frac{1}{|\mathcal{D}^l|T} \sum_{\tau \in {\mathcal{D}^l}} \sum_{t=0}^T [V_{\phi}(s_t) - \widetilde{V}_{\pi (\theta_p^l)} (s_t)] \tag{A.2}
		        \end{equation}
		        \State Store $\mathcal{D}^l$ into $\mathcal{D}^G$ in the tuple format $\langle s_t, a_t, s_{t+1} \rangle$
		    \EndFor
		    \State Output $\theta_p^k$ and $\phi_p^k$
		    \State

			\For {$m=1,2,\ldots$}
			    \State State-action pairs $\langle s, a \rangle$ in $\mathcal{D}^G$ were labeled as 0
			    \State Sample batches from $\mathcal{D}^E$ and $\mathcal{D}^G$
				\State Update $w^m$ by maximizing:
				\begin{equation}
				    w^{m+1} = \max_w \mathbb{E}_{(s, a) \sim {\mathcal{D}_E}}[log(D_w(s, a))] + \mathbb{E}_{(s, a) \sim {\mathcal{D}_G}}[1 - log D_w(s, a)] \tag{A.3}
				\end{equation}
			\EndFor
			\State Output $w^k$
			\State

			\For {$n=1,2,\ldots$}
			    \State Sample a batch from $\mathcal{D}^G$
				\State Update $\theta_{ei}^n$ and $\theta_f^n$ by minimizing the loss defined in Equation (12):
				\begin{equation}
				    \langle \theta_{ei}^{n+1}, \theta_{f}^{n+1} \rangle = \min_{\theta_{ei}, \theta_f} \mathcal{L}_{curiosity} = \min_{\theta_{ei}, \theta_f} (1 - \beta) \mathcal{L}_{ei} + \beta \mathcal{L}_f \tag{A.4}
				\end{equation}
			\EndFor
			\State Output $\theta_{ei}^k$ and $\theta_f^k$
			\State

			\State Update the reward function:
			\begin{equation}
			    \widetilde{r}^k = r_{e} + \lambda_d r_d^k + \lambda_c r_c^k \tag{A.5}
			\end{equation}
		\EndFor
		\State Output the parameters $\langle \theta_p, \phi_p, w, \theta_{ei}, \theta_f \rangle$
	\end{algorithmic} 
\end{algorithm}

\newpage
\section{Asynchronous PGfDC}

\begin{algorithm}
	\caption{Asynchronous PGfDC with PPO} 
	\begin{algorithmic}[1]
	\State Initialize policy parameters $\theta_p^0$, value function parameters $\phi_p^0$, and clip $\epsilon$
	\State Initialize discriminator parameters $w^0$, curiosity parameters $\langle \theta_{ei}^0, \theta_{f}^0 \rangle$
	\State Initialize experience replay buffer $\mathcal{D}^G$ to store generated trajectories
	\State Load expert demonstrations $\mathcal{D}^E$, state-action pairs $\langle s, a \rangle$ in $\mathcal{D}^E$ were labeled as 1
		    \State
		    \State \textbf{Process 1: PPO}
		    \Repeat
		        \State Collect set of trajectories $\mathcal{D}^g$ by running policy $\pi (\theta_p)$ in the environment
		        \State Compute state value estimates, $\widetilde{V}_{\pi (\theta_p)} (s_t)$
		        \State Compute advantage estimates, $\widetilde{A}_{\pi (\theta_p)} (s_t, a_t)$
		        \State Update $\theta_p$ by maximizing the PPO-Clip objective:
		        \begin{equation}
		            \theta_p = argmax_{\theta} \frac{1}{|\mathcal{D}^g|T} \sum_{\tau \in {\mathcal{D}^g}} \sum_{t=0}^T \min [\frac{\pi_{\theta} (s_t, a_t)}{\pi_{\theta_p} (s_t, a_t)} A^{\pi_{\theta_p}} (s_t, a_t), g \langle \epsilon, A^{\pi_{\theta_p}} (s_t, a_t) \rangle] \tag{A.6}
		        \end{equation}
		        \State Update $\phi_p$ by minimizing the mean-squared error:
		        \begin{equation}
		            \phi_p = argmin_{\phi} \frac{1}{|\mathcal{D}^g|T} \sum_{\tau \in {\mathcal{D}^g}} \sum_{t=0}^T [V_{\phi}(s_t) - \widetilde{V}_{\pi (\theta_p)} (s_t)] \tag{A.7}
		        \end{equation}
		        \State Store $\mathcal{D}^g$ into $\mathcal{D}^G$ in the tuple format $\langle s_t, a_t, s_{t+1} \rangle$
		        \State \textbf{Periodically}
		        \State $\longmapsto$ Request for the latest discriminator parameters $w$ from Process 2
		        \State $\longmapsto$ Request for the latest curiosity parameters $\langle \theta_{ei}, \theta_f \rangle$ from Process 3
		        \State $\longmapsto$ Update the reward function:
		        \begin{equation}
			        \widetilde{r} = r_{e} + \lambda_d r_d + \lambda_c r_c \tag{A.8}
			    \end{equation}
		    \Until True, output the parameters $\langle \theta_p, \phi_p, w, \theta_{ei}, \theta_f \rangle$
		    \State
		    
		    \State \textbf{Process 2: discriminator learner}
		    \Repeat
			    \State State-action pairs $\langle s, a \rangle$ in $\mathcal{D}^G$ were labeled as 0
			    \State Sample batches from $\mathcal{D}^E$ and $\mathcal{D}^G$
				\State Update $w$ by maximizing:
				\begin{equation}
				    w = \max_w \mathbb{E}_{(s, a) \sim {\mathcal{D}_E}}[log(D_w(s, a))] + \mathbb{E}_{(s, a) \sim {\mathcal{D}_G}}[1 - log D_w(s, a)] \tag{A.9}
				\end{equation}
			    \State Upon request, return the latest discriminator parameters $w$
			\Until True
			\State 
			
			\State \textbf{Process 3: curiosity learner}
			\Repeat
			    \State Sample a batch from $\mathcal{D}^G$
				\State Update $\theta_{ei}$ and $\theta_f$ by minimizing the loss defined in Equation (12):
				\begin{equation}
				    \langle \theta_{ei}, \theta_{f} \rangle = \min_{\theta_{ei}, \theta_f} \mathcal{L}_{curiosity} = \min_{\theta_{ei}, \theta_f} (1 - \beta) \mathcal{L}_{ei} + \beta \mathcal{L}_f \tag{A.10}
				\end{equation}
				\State Upon request, return the latest curiosity parameters $\langle \theta_{ei}, \theta_f \rangle$
			\Until True
	\end{algorithmic} 
\end{algorithm}

\end{document}